\documentclass[conference]{IEEEtran}
\IEEEoverridecommandlockouts
\usepackage{cite}
\usepackage{amsmath,amssymb,amsfonts}
\usepackage{subcaption}
\usepackage{tikz}
\usepackage{array}
\usetikzlibrary{positioning, shapes.geometric, arrows}
\usetikzlibrary{shapes,arrows}
\usetikzlibrary{positioning, arrows.meta}
\usepackage{algorithmic}
\usepackage{graphicx}
\usepackage{comment}
\usetikzlibrary{calc}

\usepackage{xcolor}
\usepackage{hyperref}
\hypersetup{
    colorlinks=true,
    linkcolor=blue,
    filecolor=magenta,
    urlcolor=cyan,
    pdftitle={Overleaf Example},
    pdfpagemode=FullScreen,
    }

\usepackage{textcomp}
\usepackage{xcolor}
\def\BibTeX{{\rm B\kern-.05em{\sc i\kern-.025em b}\kern-.08em
    T\kern-.1667em\lower.7ex\hbox{E}\kern-.125emX}}

\newcommand{\nix}[1]{}

\begin{document}
\title{{Advanced Arabic Alphabet Sign Language Recognition Using Transfer Learning and Transformer Models}}

\author{\IEEEauthorblockN{Mazen Balat$^\dag$,~~~ Rewaa Awaad$^\dag$,~~~ Hend Adel$^\dag$,~~~ Ahmed B. Zaky$^\dag$,~~~Salah A. Aly$^\S$ }
\IEEEauthorblockA{\textit{$^\dag$CS \& IT Dept.,}
\textit{Egypt-Japanese University of Science \& Technology,
Alexandria,  Egypt}\\
\textit{$^\S$Faculty of Computing \& Data Science,}
\textit{Badya University,
Giza, Egypt}\\
}
}

\maketitle

\begin{abstract}

This paper presents an Arabic Alphabet Sign Language recognition approach, using deep learning methods in conjunction with transfer learning and transformer-based models. We study the performance of the different variants on two publicly available datasets, namely ArSL2018 and AASL. This task will make full use of state-of-the-art CNN architectures like ResNet50, MobileNetV2, and EfficientNetB7, and the latest transformer models such as Google ViT and Microsoft Swin Transformer. These pre-trained models have been fine-tuned on the above datasets in an attempt to capture some unique features of Arabic sign language motions. Experimental results present evidence that the suggested methodology can receive a high recognition accuracy, by up to 99.6\% and 99.43\% on ArSL2018 and AASL, respectively. That is far beyond the previously reported state-of-the-art approaches. This performance opens up even more avenues for communication that may be more accessible to Arabic-speaking deaf and hard-of-hearing, and thus encourages an inclusive society.

\end{abstract}


\section{Introduction}

Sign language serves as a vital bridge between the hearing and deaf worlds \cite{othman2024sign}, with Arabic Alphabet Sign Language (ArASL) holding particular significance due to the widespread use of Arabic. The development of efficient ArASL recognition systems represents not just a technological challenge, but a crucial step towards creating more inclusive societies in Arabic-speaking regions \cite{hassan2024enhancing}. These systems have the potential to revolutionize communication, education, and social integration for the deaf and hard-of-hearing community.

Advanced Arabic Alphabets Sign Language (ArASL) recognition has significant societal impacts, including improving education, workplace integration, and public services access \cite{almubayei2024sign}. It can also improve healthcare access for deaf patients in rural areas, and provide real-time translation during emergencies \cite{abdul2024empowering}. Integrating sign language recognition technology into mainstream devices can raise awareness of deaf culture, leading to more inclusive policy-making and societal attitudes \cite{yeratziotis2023making}. Thus, accurate and efficient Arabic Sign Language recognition is a crucial step towards digital inclusivity and equal access to information and services.

Recent advancements in deep learning, especially in the domains of computer vision and sequence modeling, have opened new avenues for developing robust recognition systems. Transfer Learning \cite{Hosna2022} has emerged as a particularly promising approach, allowing researchers to leverage pre-trained models to enhance performance and generalization while reducing training time. This method shows great potential in unraveling the intricacies of ArASL and building scalable recognition systems.

In this work, we present our multifaceted contributions to the field of ArASL recognition, aiming to address these challenges and push the boundaries of what is possible in sign language technology:
\begin{enumerate}
    \item[i] We achieve superior recognition accuracy through the innovative application of transfer learning and state-of-the-art Transformer-based models.
    \item[ii] We develop a scalable ArASL recognition framework that is adaptable to other sign languages, promoting wider applicability and impact.
    \item[iii] We facilitate the creation of practical communication tools specifically designed for the Arabic-speaking deaf community, bridging the gap between technological advancement and real-world application.
\end{enumerate}

\begin{figure}[htbp]
    \centering
    \begin{tikzpicture}[node distance=0.7cm, auto,
        block/.style={rectangle, draw, fill=blue!20,
            text width=2cm, text centered, rounded corners, minimum height=0.8cm, font=\footnotesize},
        line/.style={draw, -latex'}]
        we
        \node [block] (input) {Input Sign Image};
        \node [block, right=of input] (preprocess) {Preprocessing};
        \node [block, below=of preprocess] (feature) {Feature Extraction};
        \node [block, left=of feature] (classify) {Classification};
        \node [block, below=of classify] (output) {Recognized Letter};

        \node [right=0.2cm of feature, font=\tiny] (cnn) {CNN Models};
        \node [below=0.2cm of feature, font=\tiny] (transformer) {Transformers};

        \path [line] (input) -- (preprocess);
        \path [line] (preprocess) -- (feature);
        \path [line] (feature) -- (classify);
        \path [line] (classify) -- (output);
        \path [line] (cnn) -- (feature);
        \path [line] (transformer) -- (feature);
        \draw [dashed] ($(classify.north west)+(-0.2,0.5)$) rectangle ($(feature.south east)+(1.5,-0.5)$);
    \end{tikzpicture}
    \caption{Arabic Alphabet Sign Language Recognition System}
    \label{fig:asl_recognition_system}
\end{figure}
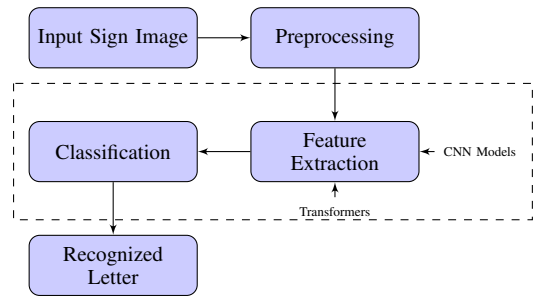

Figure \ref{fig:asl_recognition_system} illustrates the core components of our Arabic Alphabet Sign Language Recognition System. This system encompasses several key stages, including input processing, feature extraction using both CNN models and transformers, and classification, ultimately leading to the recognition of individual letters.

By harnessing these innovative techniques, our goal is to redefine the landscape of ArASL recognition. We aim to empower the Arabic-speaking deaf and hard-of-hearing community, paving the way for a more inclusive and accessible future. This initiative represents more than just a technological improvement; it embodies our commitment to create a society that is more equitable and inclusive for all its members.

The structure of this paper is given as follows: Section~\ref{sec:relatedworks} provides a comprehensive review of pertinent literature in Arabic Sign Language (ArSL) recognition. Section~\ref{sec:dataset} describes the ArSL2018 and AASL datasets utilized in our study. Our methodology, including preprocessing techniques and model architectures, is detailed in Section~\ref{sec:model}. Section~\ref{sec:metrics} presents and discusses our experimental results and performance comparisons. Finally, Section~\ref{sec:conclusion} summarizes our findings, concludes the paper, and offers insights into future research directions in this critical field.

\section{Related Work}\label{sec:relatedworks}

Contemporary methodologies for facilitating sign language communication, including human interpretation \cite{Balaha2023}, written communication \cite{app13010453}, and Automatic Speech Recognition \cite{Shashidhar2023}, while valuable, often exhibit limitations in scope and efficacy. The intricate and dynamic nature of sign languages, particularly Arabic Alphabets Sign Language (ArASL), presents significant challenges to conventional machine learning paradigms \cite{Li2021}. These traditional approaches frequently struggle to capture the nuanced gestures and diverse signing patterns characteristic of ArASL.

Recent advancements in the field have yielded promising results. A notable study proposed a sign language recognition system utilizing transfer learning techniques for Arabic alphabets, leveraging the ArSL2018 dataset. The researchers implemented preprocessing techniques to mitigate class imbalance, including image resizing and data augmentation through horizontal and vertical shifts with zooming. Employing the EfficientNetB4 model, this approach achieved a commendable testing accuracy of 95\%, demonstrating the potential of transfer learning in Arabic sign language recognition \cite{hu2022sign}.

Abdelghfar et al. introduced a deep learning approach for Qur'anic sign language recognition in 2024. Utilizing a subset of the ArSL2018 dataset, they addressed class imbalance through Random Oversampling, Synthetic Minority Over-sampling Technique, and Random Undersampling. Their QSLRS-CNN model attained impressive accuracies of 97.13\% and 97.31\% at the 100th and 200th epochs, respectively, surpassing existing models in performance \cite{abdelghfar2023model}.

El Baz et al. conducted a comprehensive study on Arabic alphabet sign language recognition using deep learning techniques and the RGB Arabic Alphabets Sign Language Dataset. Their methodology incorporated data preprocessing, including cleaning, resizing, and background removal, followed by data augmentation. The proposed architecture, comprising convolutional, pooling, dense, and dropout layers, achieved remarkable accuracies of 99.4\% in training and 97.4\% in validation \cite{el2024deep}.

Al Nabih et al. explored a Vision Transformer (ViT)-based approach for Arabic sign language recognition. By fine-tuning a pre-trained ViT model on the ArSL2018 dataset, they achieved an outstanding accuracy of 99.3\%, outperforming several recent CNN-based approaches \cite{alnabih2024arabic}.

Lahiani et al. conducted a comparative analysis of three pre-trained CNN-based architectures—InceptionV3, VGG16, and MobileNetV2—for Arabic alphabet sign language recognition. Utilizing transfer learning techniques on the ArSL2018 dataset, their study revealed superior performance with the MobileNetV2 network, achieving an accuracy of 96\% \cite{lahiani2024exploring}.

Renjith et al. proposed an innovative approach to sign language recognition by leveraging spatio-temporal features. Their method, applied to both Chinese Sign Language (CSL) and Arabic Alphabets Sign Language (ArASL), demonstrated promising results with accuracies of 90.87\% for CSL and 89.46\% for ArSL alphabet recognition \cite{renjith2024sign}.

The literature reveals the efficacy of various pre-trained models in recognition tasks. Architectures such as ResNet50 \cite{he2016deep}, MobileNetV2 \cite{mobilenetv22018}, and EfficientNetB7 \cite{Tan2019EfficientNetRM} have demonstrated remarkable results in image classification. More recently, transformer networks like Google's Vision Transformer (ViT) \cite{wu2020visual} and Microsoft's Swin Transformer \cite{DBLP:journals/corr/abs-2103-14030} have revolutionized computer vision through self-attention mechanisms and hierarchical approaches. The superior performance of these pre-trained models, when fine-tuned for sign language recognition tasks, underscores their adaptability and effectiveness in this domain.

\section{Datasets}\label{sec:dataset}

In this study, we utilize two datasets for Arabic alphabet sign language recognition: the Arabic Sign Language ArSL2018 dataset and RGB Arabic Alphabets Sign Language dataset (ArASL).

\subsection{Arabic Alphabets Sign Language Dataset (ArASL2018)}

\begin{figure}[h]
\centering

\includegraphics[width=0.15\textwidth]{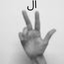}
\includegraphics[width=0.15\textwidth]{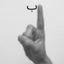}
\includegraphics[width=0.15\textwidth]{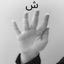}

\caption{Examples of images from the ArSL2018 dataset }
\label{fig:ArSL2018_images}
\end{figure}

A significant contribution to Arabic Sign Language (ArSL) recognition research was made by Latif et al. \cite{latif2018arabic} with the introduction of the ArSL2018 dataset. This comprehensive dataset consists of 54,049 grayscale images (64x64 pixels) representing 32 Arabic sign language signs and alphabets. The images were collected from 40 participants of various age groups in Al Khobar, Saudi Arabia, using an iPhone 6S camera. The dataset includes variations in lighting, angles, and backgrounds to enhance its robustness. Examples of these images can be seen in Figure \ref{fig:ArSL2018_images}. ArSL2018 is notable for being one of the first large, fully-labeled datasets for Arabic Sign Language, making it a valuable resource for researchers developing machine learning and computer vision applications for the deaf and hard of hearing community. The authors reported achieving high accuracy in their initial experiments, establishing a benchmark for future research. This dataset addresses a crucial need in the field, as it enables faster development and prototyping of assistive technology applications specific to Arabic sign language.

\subsection{RGB Arabic Alphabets Sign Language Dataset (ArASL)}

\begin{figure}[h]
\centering
\includegraphics[width=0.15\textwidth]{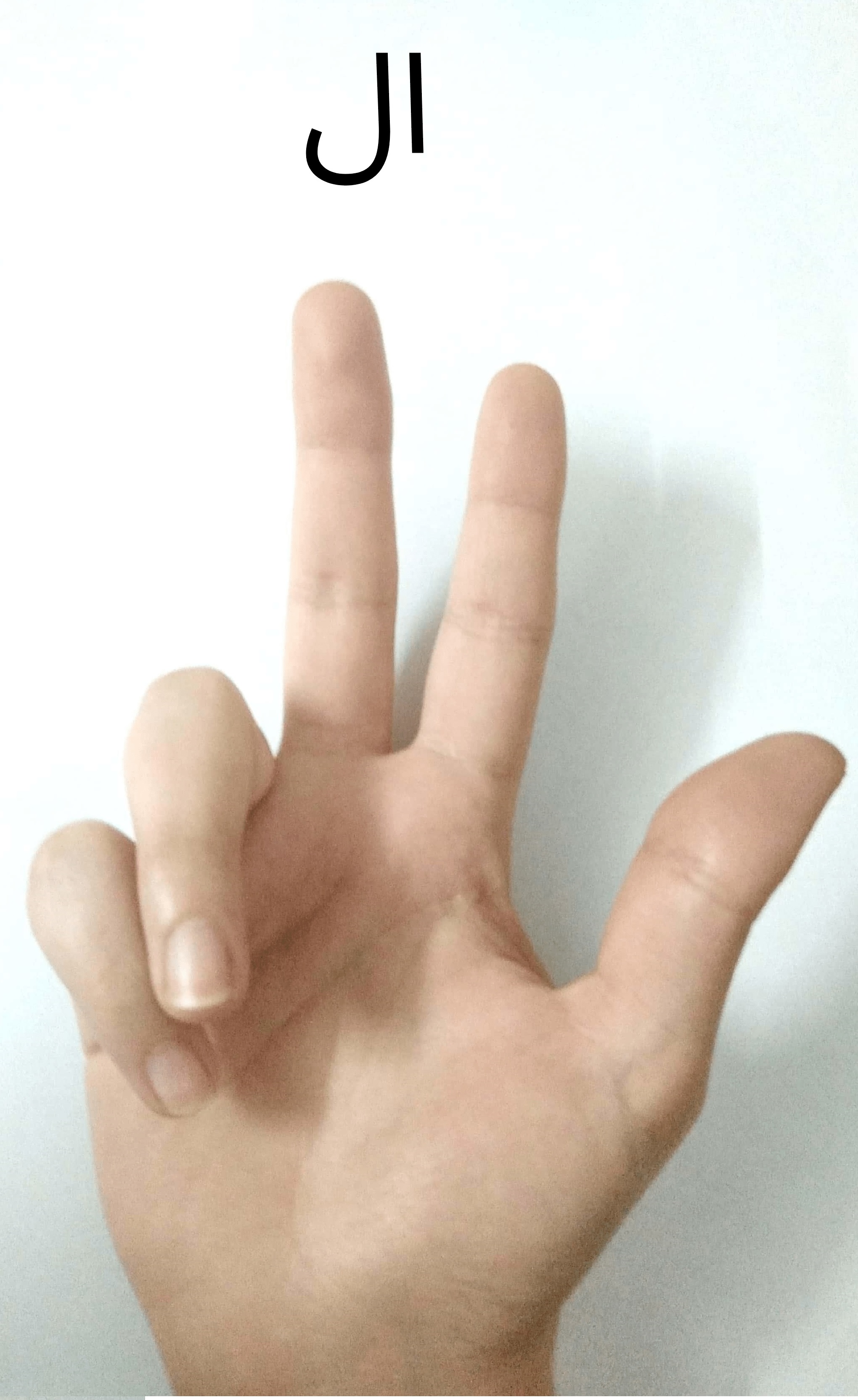}
\includegraphics[width=0.15\textwidth]{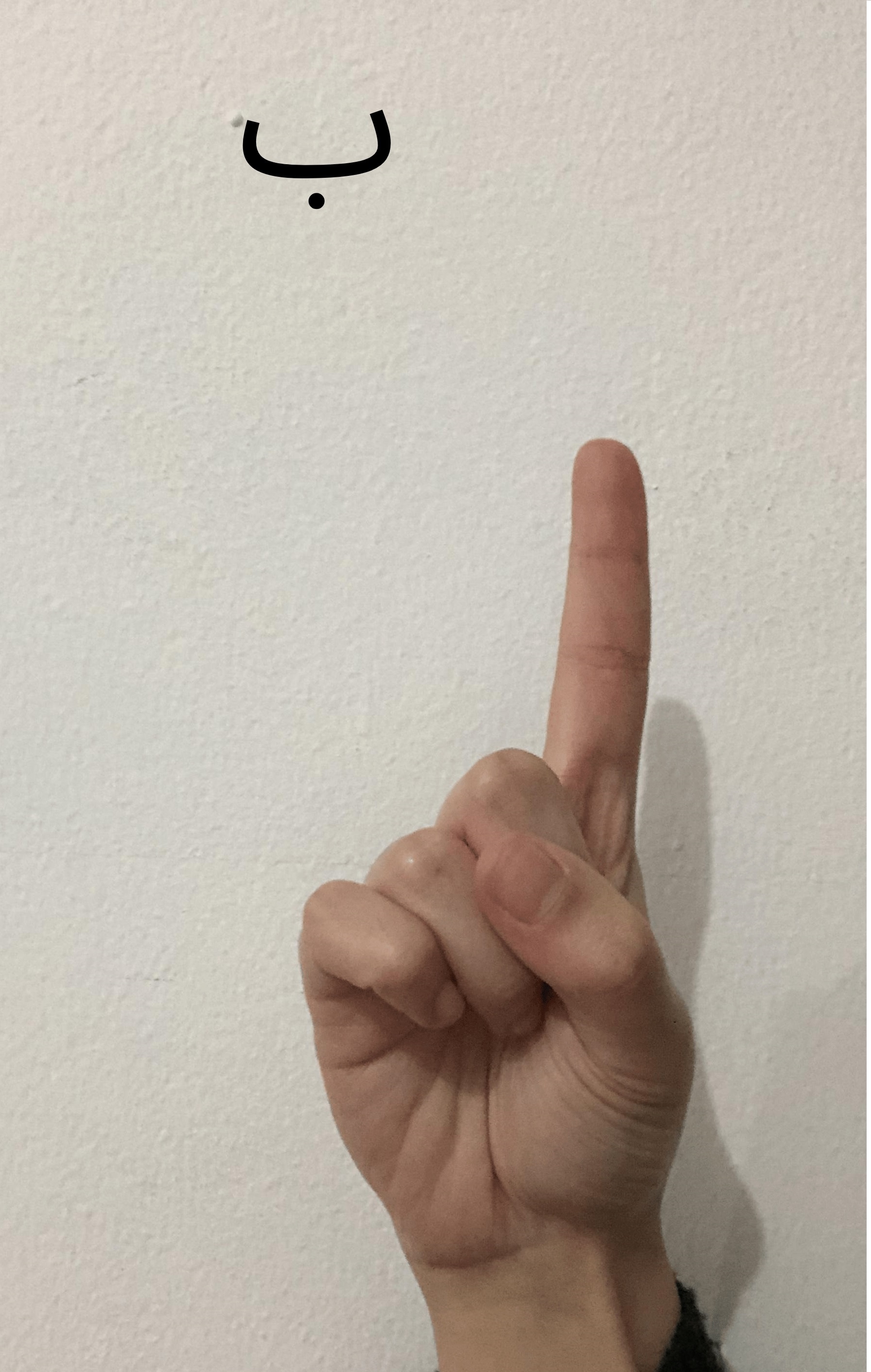}
\includegraphics[width=0.15\textwidth]{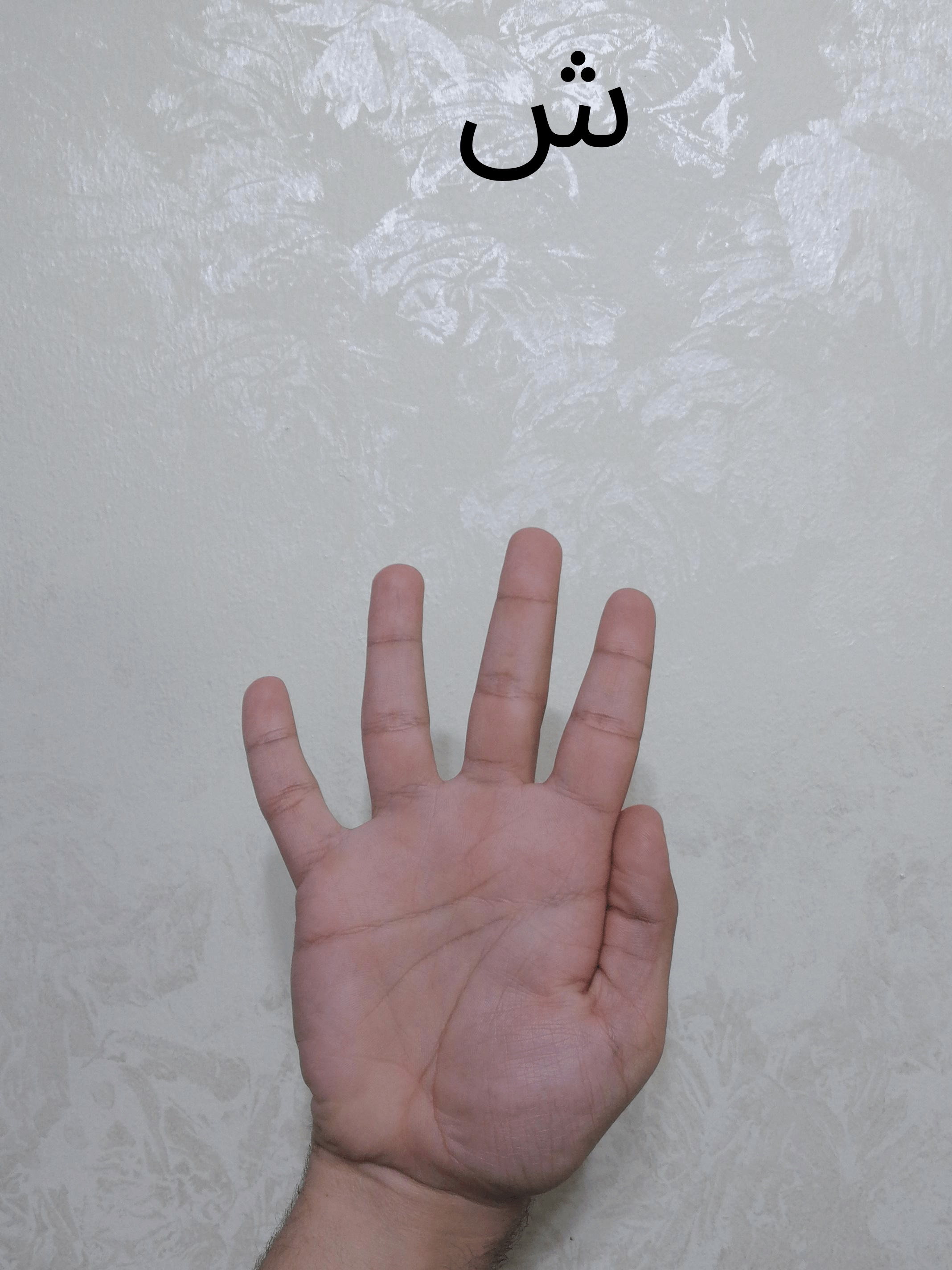}
\caption{Examples of images from the AASL dataset}
\label{fig:AASL_images}
\end{figure}

The RGB Arabic Alphabets Sign Language (AASL) dataset \cite{https://doi.org/10.48550/arxiv.2301.11932} consists of 7,857 labeled RGB images, representing 31 Arabic sign language alphabets. Collected from over 200 participants using various types of cameras (webcams, digital cameras, and phone cameras), the dataset captures a range of conditions including different lighting, backgrounds, and orientations. This diversity enhances its robustness for real-world applications. Experts validated and filtered the images to ensure high quality, making AASL an essential dataset for developing accurate Arabic sign language classification models. Examples of images from the dataset are displayed in Figure \ref{fig:AASL_images}.

\section{Methodology}
\label{sec:model}

The methodology for Arabic Alphabet Sign Language recognition consists of three key steps: \textbf{data pre-processing}, \textbf{model selection with transfer learning}, and \textbf{model evaluation}. Each stage plays a crucial role in ensuring the system's accuracy and reliability. The following sections will detail these procedures to provide a clear understanding of the research approach.

\subsection{Data Preprocessing}

Data preprocessing mainly helps in refining the dataset so that the deep learning models may be best trained. A few key tasks comprise the standardization of input data or enhancing its quality in the preprocessing pipeline.

Class imbalance is one of the major issues to be tackled in this step. Few of the Arabic alphabet signs have much more samples than the rest, and hence using them would make predictions biased. Therefore, methods like under-sampling of majority classes and over-sampling of minority classes are put into practice. Each of the 28 Arabic alphabet signs must be represented decently while preparation of the dataset.

After balancing classes, these images undergo grayscale conversion to reduce dimensionality, allowing a greater focus on the shape and texture of hand gestures. Since color information does not play an important role in recognizing hand signs, a gray-scale conversion of the problem at hand can help to simplify it further and reduce computational complexity.

The images are then uniformly resized to 224x224 pixels-a requirement to be compatible with some of these pre-trained models, such as ResNet50, VGG16, and MobileNetV2. This standardization of the image dimensions was performed for easy training of the model and to make all the input data uniform.

Normalization of the pixel values scales them between a range of [0, 1]. This helps reduce light variations and adds more robustness to the model for different lighting conditions during inference.

Lastly, the data is divided into training, validation, and test subsets. A standard split of 70\% for training, 15\% for validation, and 15\% for testing is done. This splitting ensures that the model is tested on unseen data for hyperparameter tuning and allows for a more realistic estimate of performances.

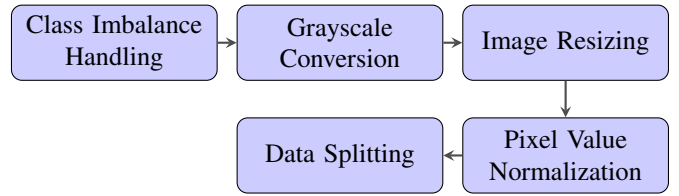
\begin{figure}[htbp]
    \centering
    \tikzstyle{block} = [rectangle, draw, fill=blue!20, rounded corners, text width=2.5cm, text centered, minimum height=1cm]
    \tikzstyle{arrow} = [thick,->,>=stealth, color=black!70]

    \begin{tikzpicture}[node distance=3cm]

    \node (imbalance) [block] {Class Imbalance Handling};
    \node (grayscale) [block, right of=imbalance] {Grayscale Conversion};
    \node (resize) [block, right of=grayscale] {Image Resizing};
    \node (normalize) [block, below of=resize, node distance=1.5cm] {Pixel Value Normalization};
    \node (split) [block, left of=normalize] {Data Splitting};

    \draw [arrow] (imbalance) -- (grayscale);
    \draw [arrow] (grayscale) -- (resize);
    \draw [arrow] (resize) -- (normalize);
    \draw [arrow] (normalize) -- (split);

    \end{tikzpicture}
    \caption{The preprocessing steps applied to the dataset}
    \label{fig:preproc_meth}
\end{figure}

As shown in Figure \ref{fig:preproc_meth}, these steps, including class imbalance handling, grayscale conversion, image resizing, pixel value normalization, and data splitting, form the core of the preprocessing pipeline, leading to improved data quality and standardized inputs.

\subsection{Model Selection and Transfer Learning}

Our approach to Arabic Alphabet Sign Language recognition utilizes transfer learning with both Convolutional Neural Network (CNN) architectures and transformer-based models. We selected five pre-trained models to leverage their feature extraction capabilities learned from large-scale datasets: ResNet50, MobileNetV2, EfficientNetB7, Google Vision Transformer (ViT), and Microsoft Swin Transformer.

For the CNN-based models (ResNet50, MobileNetV2, and EfficientNetB7), we employ a fine-tuning strategy. The early layers of these models are frozen to retain general features learned from ImageNet, while the final classification layers are replaced with new layers configured for our 28-class Arabic alphabet recognition task. Only these new layers and a few preceding layers are left unfrozen, allowing fine-tuning on the specific characteristics of our dataset.

For the transformer-based models (Google ViT and Microsoft Swin), we adopt a similar fine-tuning approach. These models, originally trained on large image datasets, are adapted to our specific task by modifying their classification heads while keeping the core transformer blocks mostly frozen.

All models are trained with a batch size of 32, using the Adam optimizer with an initial learning rate of 0.001. We employ the cross-entropy loss function to compute the difference between predicted and actual class labels. A StepLR scheduler is implemented to decay the learning rate by a factor of 0.1 every 10 epochs, aiding in better convergence. To prevent overfitting, we implement early stopping: training halts if the validation accuracy does not improve for 5 consecutive epochs.

\begin{figure}[htbp]
    \centering
    \resizebox{0.45\textwidth}{!}{ 
        \begin{tikzpicture}[node distance=0.8cm and 0.8cm] 
            \tikzstyle{block} = [rectangle, draw, fill=blue!20, rounded corners, minimum width=2.2cm, minimum height=1cm, text centered, text width=2.2cm]
            \tikzstyle{arrow} = [thick, ->, >=stealth, color=black!70]

            \node[block] (premodel) {\parbox{2 cm}{Pre-trained \\ Backbone}};
            \node[block, right=of premodel] (featmaps) {\parbox{2 cm}{Feature Maps \\ (512x7x7)}};
            \node[block, below=of featmaps] (fc) {\parbox{2.2 cm}{New Fully \\ Connected Layers}};
            \node[block, left=of fc] (out) {\parbox{2 cm}{Output \\ (32 \& 31 classes)}};

            \draw[arrow] (premodel) -- (featmaps);
            \draw[arrow] (featmaps) -- (fc);
            \draw[arrow] (fc) -- (out);

            \node[left=of premodel] (in) {\parbox{2 cm}{Input Image \\ (224x224x3)}};

            \draw[arrow] (in) -- (premodel);

        \end{tikzpicture}
    }
    \caption{The transfer learning and fine-tuning process for Arabic alphabet sign language recognition}
    \label{fig:trans_meth}
\end{figure}
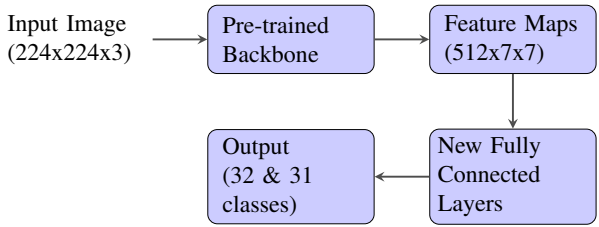

Figure \ref{fig:trans_meth} illustrates the transfer learning process, which is applicable to both CNN and transformer architectures. The input image is passed through the pre-trained backbone (ResNet50, MobileNetV2, EfficientNetB7, Google ViT, or Microsoft Swin), which extracts feature maps. These feature maps are then fed into new fully connected layers specifically designed for Arabic alphabet classification, leading to the final output predictions.

This approach allows us to leverage the strengths of both CNN and transformer architectures, potentially capturing different aspects of the sign language images and leading to robust recognition performance.

\subsection{Model Evaluation}

The performance of our models is evaluated using accuracy as the primary metric \cite{Rainio2024}. Accuracy provides a straightforward measure of the model's overall performance in classifying Arabic alphabet signs.

Accuracy measures the proportion of correctly classified instances out of the total instances in the dataset. It is mathematically defined as:

\begin{equation}
\text{Accuracy} = \frac{TP + TN}{TP + TN + FP + FN}
\end{equation}

While accuracy offers a clear indication of overall performance, it's important to note that in cases of class imbalance, it may not fully capture the model's effectiveness across all classes. However, given the balanced nature of our datasets (ArASL2018 and AASL), accuracy serves as an appropriate and sufficient metric for our evaluation.

We report accuracy for training, validation, and test sets to provide a comprehensive view of each model's performance and to assess potential overfitting or underfitting issues.

\section{Results}\label{sec:metrics}

In this section, we present the results of our experiments on Arabic alphabet sign language recognition using fine-tuned CNN and transformer models. We evaluate the performance of these models on test datasets and compare their accuracy, training time, and overall efficiency to state-of-the-art methods.

\subsection{ArASL2018 Dataset}

Table \ref{tab:transfer_learning_results_arasl} and Table \ref{tab:transformer_results_arasl} summarize the performance of both transfer learning and transformer-based models on the ArASL2018 dataset. The models were evaluated based on training, validation, and test accuracy, as well as the time taken for training.

\begin{table}[ht]
\centering
\caption{Transfer Learning Results on ArASL2018 Dataset}
\label{tab:transfer_learning_results_arasl}
\begin{tabular}{lcccc}
\hline
Model & Train Acc & Val Acc & Test Acc & Train Time \\
\hline
Resnet50 & 99.91\% & 99.43\% & 99.30\% & 60.94 minutes \\
MobileNetV2 & \textbf{99.92\%} & \textbf{99.45\%} & 99.48\% & \textbf{26.52 minutes} \\
EfficientNetB7 & 99.91\% & 99.33\% & \textbf{99.60\%} & 201.75 minutes \\
\hline
\end{tabular}
\end{table}

\begin{table}[ht]
\centering
\caption{Transformer-based Results on ArASL2018 Dataset}
\label{tab:transformer_results_arasl}
\begin{tabular}{lcccc}
\hline
Model & Train Acc & Val Acc & Test Acc & Train Time \\
\hline
Google ViT & \textbf{99.91\%} & 99.18\% & 99.38\% & \textbf{133.27 minutes} \\
Microsoft Swin & 99.60\% & \textbf{99.50\%} & \textbf{99.60\%} & 580.50 minutes \\
\hline
\end{tabular}
\end{table}

The results indicate that transfer learning models generally outperform transformer-based models in terms of training efficiency, with MobileNetV2 achieving the fastest training time at 26.52 minutes while maintaining a competitive test accuracy of 99.48\%. However, when test accuracy is prioritized over speed, transformer models like Microsoft Swin are superior, achieving a test accuracy of 99.6\%, albeit with a significantly longer training time of 580.50 minutes. This suggests that while transfer learning models are more computationally efficient, transformer models may offer slight improvements in accuracy for more demanding applications.

\begin{figure}[ht]
\centering
\includegraphics[width=0.45\textwidth,height=0.3\textheight]{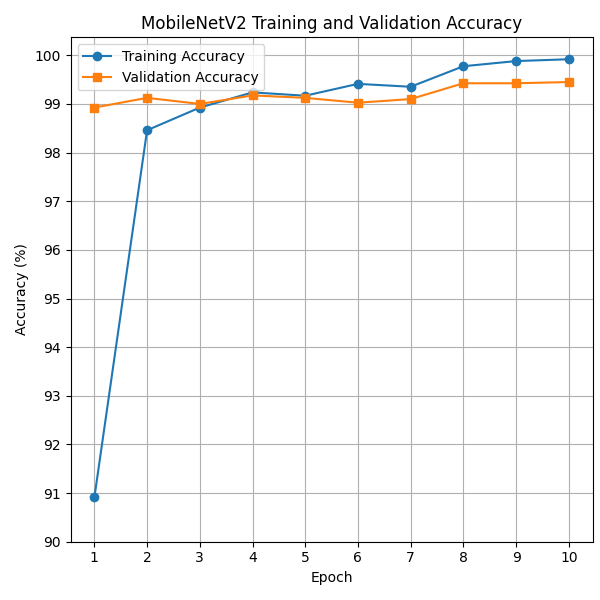}
\caption{Training and validation accuracy for MobileNetV2 on the ArASL2018 dataset.}
\label{fig:mobnet_acc_graph}
\end{figure}

\begin{figure}[ht]
\centering
\includegraphics[width=0.45\textwidth,height=0.3\textheight]{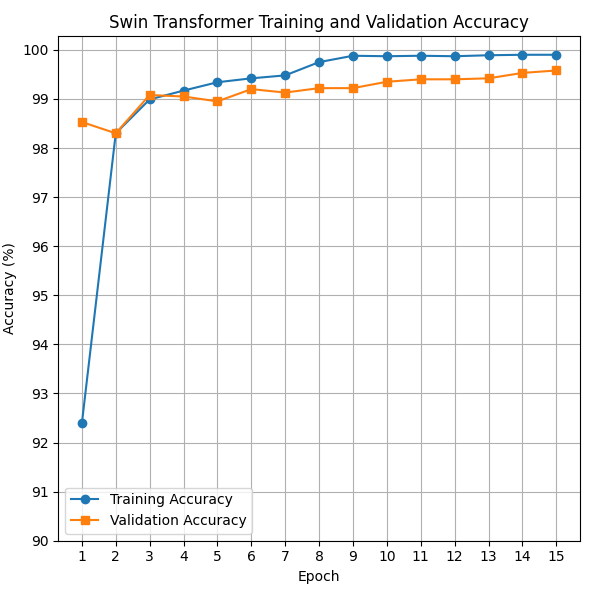}
\caption{Training and validation accuracy for Microsoft Swin on the ArASL2018 dataset.}
\label{fig:swin_acc_graph}
\end{figure}

Figure \ref{fig:mobnet_acc_graph} and Figure \ref{fig:swin_acc_graph} illustrate the training and validation accuracy for MobileNetV2 and Microsoft Swin on the ArASL2018 dataset, respectively. MobileNetV2 shows faster convergence with fewer fluctuations compared to the transformer-based Microsoft Swin model. The stability of MobileNetV2 during training is particularly noteworthy, while Swin demonstrates a more gradual and fluctuating convergence pattern.

\subsection{AASL Dataset}

Similar trends were observed on the AASL dataset, as summarized in Table \ref{tab:transfer_learning_results_aasl} and Table \ref{tab:transformer_results_aasl}. Here, MobileNetV2 once again emerged as the most efficient model in terms of training time, completing the process in 146.02 minutes while achieving a test accuracy of 99.00\%. On the other hand, the Google ViT transformer model achieved the highest test accuracy of 99.43\%, albeit at the cost of a slightly longer training time.

\begin{table}[ht]
\centering
\caption{Transfer Learning Results on AASL Dataset}
\label{tab:transfer_learning_results_aasl}
\begin{tabular}{lcccc}
\hline
Model & Train Acc & Val Acc & Test Acc & Train Time \\
\hline
Resnet50 & \textbf{100.00\%} & 98.57\% & 98.57\% & 159.68 minutes \\
MobileNetV2 & 99.96\% & 98.71\% & \textbf{99.00\%} & \textbf{146.02 minutes} \\
EfficientNetB7 & 99.93\% & \textbf{99.28\%} & 98.89\% & 168.26 minutes \\
\hline
\end{tabular}
\end{table}

\begin{table}[ht]
\centering
\caption{Transformer-based Results on AASL Dataset}
\label{tab:transformer_results_aasl}
\begin{tabular}{lcccc}
\hline
Model & Train Acc & Val Acc & Test Acc & Train Time \\
\hline
Google ViT & \textbf{100\%} & \textbf{99.43\%} & \textbf{99.43\%} & 149.19 minutes \\
Microsoft Swin & 99.62\% & 98.28\% & 98.43\% & \textbf{136.78 minutes} \\
\hline
\end{tabular}
\end{table}

This dataset further highlights the trade-off between accuracy and computational efficiency. MobileNetV2, while offering strong performance with minimal training time, does not surpass the accuracy of transformer-based models such as Google ViT, which consistently demonstrates superior test accuracy across datasets. However, the difference in training times between these two models on the AASL dataset is minimal, suggesting that transformer models may offer a viable alternative for applications where training time is less of a concern.

\begin{figure}[ht]
\centering
\includegraphics[width=0.48\textwidth,height=0.32\textheight]{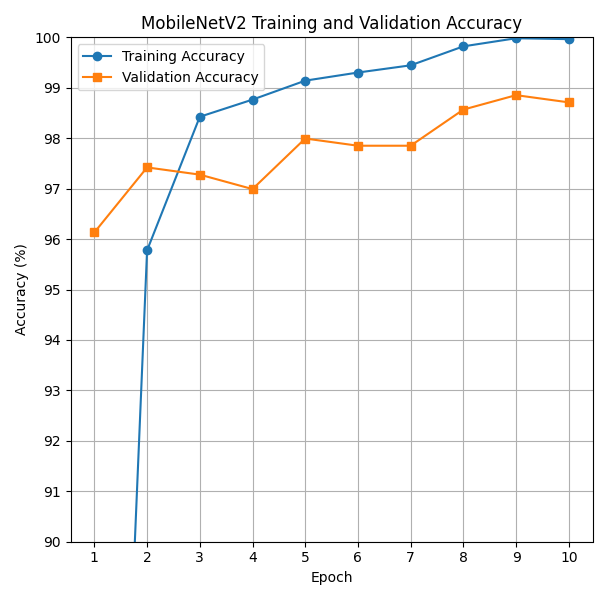}
\caption{Training and validation accuracy for MobileNetV2 on the AASL dataset.}
\label{fig:mobnet_rgb_acc_graph}
\end{figure}

\begin{figure}[ht]
\centering
\includegraphics[width=0.48\textwidth,height=0.32\textheight]{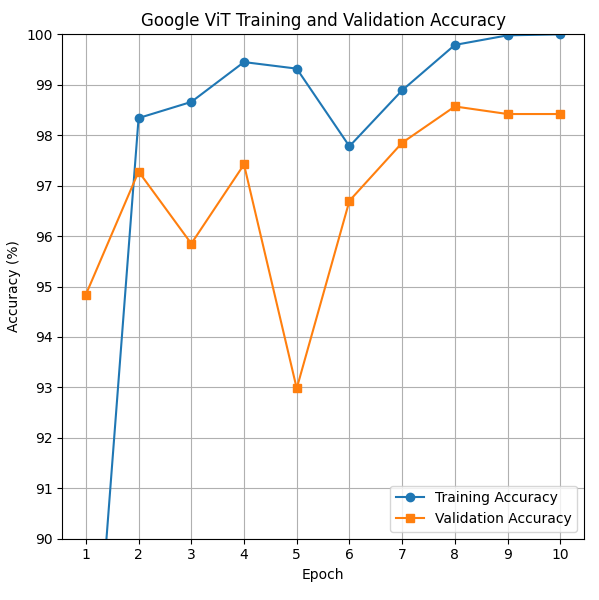}
\caption{Training and validation accuracy for Google ViT on the AASL dataset.}
\label{fig:vit_rgb_acc_graph}
\end{figure}

Figure \ref{fig:mobnet_rgb_acc_graph} and Figure \ref{fig:vit_rgb_acc_graph} compare the training and validation accuracy of MobileNetV2 and Google ViT on the AASL dataset. While both models show efficient convergence, Google ViT consistently maintains a higher validation accuracy throughout the training process, indicating better generalization compared to MobileNetV2.

\subsection{Comparison with Other Studies}

Our approach outperforms several state-of-the-art models from other studies, as outlined in Table \ref{tab:comparison_with_others}. On the ArASL2018 dataset, our Microsoft Swin model surpasses the performance of Hu et al. \cite{hu2022sign}, Abdelghfar et al. \cite{abdelghfar2023model}, and Alnabih et al. \cite{alnabih2024arabic}, achieving an impressive test accuracy of 99.60\%. Similarly, on the AASL dataset, our Google ViT model reaches a test accuracy of 99.43\%, improving upon the previous work of El-Sayed et al. \cite{el2024deep} and Renjith et al. \cite{renjith2024sign}.

\begin{table}[ht]
\centering
\caption{Comparison with Other Studies on ArASL2018 and AASL Datasets}
\label{tab:comparison_with_others}
\begin{tabular}{lcccc}
\hline
\textbf{Study} & \textbf{Dataset} & \textbf{Test Accuracy} \\
\hline
Hu et al. \cite{hu2022sign} & ArASL2018 & 94.95\% \\
Abdelghfar et al. \cite{abdelghfar2023model} & ArASL2018 & 97.31\% \\
Alnabih et al. \cite{alnabih2024arabic} & ArASL2018 & 99.30\% \\
\textbf{Our Approach (Microsoft Swin)} & ArASL2018 & \textbf{99.60\%}\\
\hline
El-Sayed et al. \cite{el2024deep} & AASL & 97.40\%\\
Renjith et al. \cite{renjith2024sign} & AASL & 89.46\% \\
\textbf{Our Approach (Google ViT)} & AASL & \textbf{99.43\%}\\
\hline
\end{tabular}
\end{table}

These improvements reflect the robustness and scalability of transformer-based architectures, particularly in their ability to generalize across multiple datasets, even when confronted with highly similar gesture patterns. The gains in accuracy observed in our approach suggest that transformers are well-suited for complex tasks such as sign language recognition, where subtle variations in gesture can have significant implications for classification performance.

\section{Conclusion}\label{sec:conclusion}

In this study, we presented an Arabic Alphabet Sign Language recognition approach using transfer learning with a transformer-based model, achieving state-of-the-art results with 99.6\% test accuracy on ArASL2018 and 99.43\% on the AASL dataset. Our approach demonstrates significant potential for enhancing communication technologies for the Arabic-speaking deaf and hard-of-hearing community. Future research could focus on implementing real-time translation, extending the method to full sentence recognition, and improving model robustness across diverse signing styles. Additionally, optimizing transformer models for resource-constrained devices would facilitate deployment in real-world applications. Expanding datasets and supporting multilingual sign language recognition could also broaden the system's impact, making it a valuable tool in assistive communication.

\bibliographystyle{IEEEtran}
\bibliography{cite}

\end{document}